\documentclass[a4paper]{article}

\usepackage{INTERSPEECH2018}
\usepackage{multirow}
\usepackage[hidelinks]{hyperref}

\DeclareMathOperator{\dec}{dec}
\DeclareMathOperator{\deca}{dec^1}
\DeclareMathOperator{\decb}{dec^2}
\DeclareMathOperator{\enca}{enc^1}
\DeclareMathOperator{\encb}{enc^2}

\DeclareMathOperator*{\softmax}{softmax}

\newcommand{\ascal}{\alpha}

\newcommand{\xvec}{\mathbf{x}}
\newcommand{\avec}{\mathbf{\alpha}}

\newcommand{\hvec}{\mathbf{h}}
\newcommand{\vvec}{\mathbf{v}}
\newcommand{\cvec}{\mathbf{c}}
\newcommand{\svec}{\mathbf{s}}
\newcommand{\yvec}{\mathbf{y}}
\newcommand{\step}[1]{_{#1}}
\newcommand{\fst}{^1}
\newcommand{\snd}{^2}

\usepackage{rotating}

\usepackage{tikz}
\usetikzlibrary{shapes.geometric}
\usetikzlibrary{patterns}
\usepackage{pgfplots}

\title{Leveraging translations for speech transcription in low-resource settings}
\name{Antonis Anastasopoulos, David Chiang\thanks{This work was generously supported by NSF Award 1464553.}}
\address{
  Department of Computer Science and Engineering\\
  University of Notre Dame, IN, USA}
\email{aanastas@nd.edu, dchiang@nd.edu}

\begin{document}

\maketitle

\begin{abstract}
Recently proposed data collection frameworks for endangered language documentation aim not only to collect speech in the language of interest, but also to collect translations into a high-resource language that will render the collected resource interpretable. We focus on this scenario and explore whether we can improve transcription quality under these extremely low-resource settings with the assistance of text translations. We present a neural multi-source model and evaluate several variations of it on three low-resource datasets. We find that our multi-source model with shared attention outperforms the baselines, reducing transcription character error rate by up to $12.3\%$.
\end{abstract}


\section{Introduction}

For many low-resource and endangered languages, speech data is easier to obtain than textual data. Oral tradition has historically been the main medium for passing cultural knowledge from one generation to the next, and an estimated ~43\% of the world's languages is still unwritten~\cite{lewis2009ethnologue}. 
Traditionally, documentary records of endangered languages are created by highly trained linguists in the field. However, modern technology has the potential to enable creation of much larger-scale resources. Recently proposed frameworks~\cite{bird-EtAl:2014:Coling} propose collection of bilingual audio, rendering the resource interpretable through translations.
New technologies have been developed to facilitate collection of spoken translations~\cite{bird-EtAl:2014:W14-22} along with speech in an endangered language, and there already exist recent examples of parallel speech collection efforts focused on endangered languages~\cite{blachon2016parallel,adda2016breaking}. The translation is usually in a high-resource language that functions as a \textit{lingua franca} of the area, as it is common for members of an endangered-language community to be bilingual.

Since speech transcription is a costly and slow process, automatically producing  transcriptions has the potential to significantly speed up documentation.  We focus on this language documentation scenario and explore methods that learn from a small number of transcribed speech utterances along with their translations. 
We use the neural attentional model~\cite{bahdanau2015} and experiment with extensions that take both speech utterances and their translations as input sources. We assume that the translations are in a high-resource language that can be automatically transcribed; therefore, in our experiments, the translation input is text instead of speech. We also explore different parameter-sharing methods across the attention mechanisms.

We experiment on three diverse low-resource language pairs. One is Ainu, a severely endangered language, with translations in English. We also experiment on a recently collected speech corpus of Mboshi~\cite{godard2017very}, with translations in French. Lastly, we evaluate our models on Spanish-English, using the CALLHOME dataset.

Our proposed multi-source model that employs a shared attention mechanism outperforms the baselines in almost all cases. In Mboshi, we find that our model reduces character error rates (CER) by~$1.2$ points. In Spanish, we observe a reduction of~$4.6$ points in CER over the strongest baseline, and more than~$14.4$ points over a speech-only baseline. In Ainu, although our multi-source model doesn't reduce the overall CER, we show that it actually is beneficial in the cases where the single-source speech transcription model has greatest difficulty.

\section{Model}

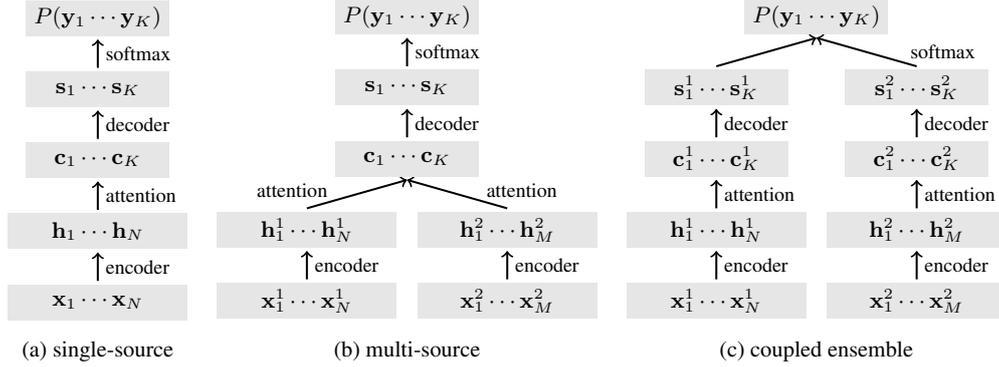
\begin{figure*}
\tikzset{seq/.style={draw=none,fill=gray!20}}
\tikzset{layer/.style={->,thick}}
\tikzset{label/.style={anchor=west,font={\footnotesize}}}
\tikzset{seqlabel/.style={font={\small}}}
\newcommand{\encoder}[2]{
\draw[seq] (-1.25,-0.25) rectangle (1.25,0.25);
\node[seqlabel] at (0,0) 
{$\xvec\step{1}#1 \cdots \xvec\step{#2}#1$};
\draw[layer] (0,0.3) -- (0,0.7);
\node[seqlabel] at (0,0.5) [label] {encoder};
\draw[seq] (-1.25,0.75) rectangle (1.25,1.25);
\node[seqlabel] at (0,1) 
{$\hvec\step{1}#1 \cdots \hvec\step{#2}#1$};
}
\newcommand{\decoder}[1]{
\draw[seq] (-1,1.75) rectangle (1,2.25);
\node[seqlabel] at (0,2) 
{$\cvec\step{1}#1 \cdots \cvec\step{K#1}#1$};
\draw[layer] (0,2.3) -- (0,2.7);
\node[seqlabel] at (0,2.5) [label] {decoder};
\draw[seq] (-1,2.75) rectangle (1,3.25);
\node[seqlabel] at (0,3) 
{$\svec\step{1}#1 \cdots \svec\step{K#1}#1$};
\draw[layer] (0,3.3) -- (0,3.7);
\node[seqlabel] at (0,3.5) [label] {softmax};
\draw[seq] (-1,3.75) rectangle (1,4.25);
\node[seqlabel] at (0,4) 
{$P(\yvec\step{1}#1 \cdots \yvec\step{K#1}#1)$};
}
\newcommand{\ensdecoder}[1]{
\draw[seq] (-1,1.75) rectangle (1,2.25);
\node[seqlabel] at (0,2) 
{$\cvec\step{1}#1 \cdots \cvec\step{K}#1$};
\draw[layer] (0,2.3) -- (0,2.7);
\node[seqlabel] at (0,2.5) [label] {decoder};
\draw[seq] (-1,2.75) rectangle (1,3.25);
\node[seqlabel] at (0,3) 
{$\svec\step{1}#1 \cdots \svec\step{K}#1$};
}
\begin{center}
\resizebox{0.8\hsize}{!}{
\begin{tabular}{ccc}
\begin{tikzpicture}
\encoder{}{N}
\draw[layer] (0,1.3) -- (0,1.7);
\node at (0,1.5) [label] {attention};
\decoder{}
\end{tikzpicture}
&
\begin{tikzpicture}
\begin{scope}[xshift=-1.4cm]
\encoder{^1}{N}
\end{scope}
\begin{scope}[xshift=1.4cm]
\encoder{^2}{M}
\end{scope}
\draw[layer] (-1.4,1.3) -- (0,1.7);
\draw[layer] (1.4,1.3) -- (0,1.7);
\node at (-1,1.55) [label,anchor=east] {attention};
\node at (1,1.55) [label] {attention};
\decoder{}
\end{tikzpicture}
&
\begin{tikzpicture}
\begin{scope}[xshift=-1.4cm]
\encoder{^1}{N}
\draw[layer] (0,1.3) -- (0,1.7);
\node at (0,1.5) [label] {attention};
\ensdecoder{^1}{K}
\end{scope}
\begin{scope}[xshift=1.4cm]
\encoder{^2}{M}
\draw[layer] (0,1.3) -- (0,1.7);
\node at (0,1.5) [label] {attention};
\ensdecoder{^2}{K}
\end{scope}
\draw[layer] (-1.4,3.3) -- (0,3.7);
\draw[layer] (1.4,3.3) -- (0,3.7);
\node[seqlabel] at (1.2,3.5) [label] {softmax};
\draw[seq] (-1,3.75) rectangle (1,4.25);
\node[seqlabel] at (0,4) 
{$P(\yvec\step{1} \cdots \yvec\step{K})$};
\end{tikzpicture}
\\[1ex]
(a) single-source & (b) multi-source & (c) coupled ensemble \\ 
\end{tabular}%
}
\end{center}
\caption{Source-side variations on the standard attentional model. In the standard \emph{single-source} model, the decoder attends to a single encoder's states. In our proposed \emph{multisource} setup, we have two input sequences encoded by two different encoders, and attention mechanisms provide two context to the decoder. Note that for clarity's sake there are dependencies not shown.}
\label{fig:multisourcemodels}
\end{figure*}

Our models are based on a sequence-to-sequence model with attention~\cite{bahdanau2015}. In general, this type of model is composed of three parts: a recurrent encoder, the attention, and a recurrent decoder (see Figure~\ref{fig:multisourcemodels}a).\footnote{For simplicity, we have assumed only a single layer for both the encoder and decoder. It is possible to use multiple stacked RNNs; typically, the output of the encoder(s) and decoder(s) ($\cvec\step{n}$ and $P(\yvec\step{k})$, respectively) would be computed from the top layer only.}

Let $\mathbf{X^1}=\xvec\fst\step{1}\ldots\xvec\fst\step{N}$ be a sequence of speech frames, $\mathbf{X}\snd = \xvec\snd\step{1}\ldots\xvec\snd\step{M}$ a sequence of translation characters, and $\mathbf{Y} = \yvec\step{1}\ldots\yvec\step{K}$ be a sequence of the target characters of the transcription. A \textit{single-source} speech recognition model attempts to model $P(\mathbf{Y}\mid\mathbf{X}\fst)$, while a \textit{single-source} translation model would model $P(\mathbf{Y}\mid\mathbf{X}\snd)$.

A \textit{multi-source} model can jointly model $P(\mathbf{Y}\mid\mathbf{X}\fst,\mathbf{X}\snd)$, and thus we need two encoders (see Figure~\ref{fig:multisourcemodels}b). One encoder transforms the input sequence of speech frames $\xvec\fst\step{1}\ldots\xvec\fst\step{N}$ into a sequence of input states $\hvec\fst\step{1}\ldots\hvec\fst\step{N}$:
\begin{align}
    \hvec\fst\step{n} &= \enca(\hvec\fst\step{n-1}, \xvec\fst\step{n})
\end{align}
\noindent and the second encoder transforms the translation character sequence $\xvec\snd\step{1}\ldots\xvec\snd\step{M}$ into another sequence of input states~$\hvec\snd\step{1}\ldots\hvec\snd\step{M}$:
\begin{align}
    \hvec\snd\step{m} &= \encb(\hvec\snd\step{m-1}, \xvec\snd\step{m}).    
\end{align}
\noindent An attention mechanism transforms the two sequences of input states into a sequence of \emph{concatenated context vectors} via two matrices of \emph{attention weights}:
\begin{align}
    \cvec\step{k} &= \begin{bmatrix} \sum_n \ascal\fst_{kn} \hvec\fst\step{n} & \sum_m \ascal\snd_{km} \hvec\snd\step{m} \end{bmatrix}.
\end{align}
\noindent Finally, the decoder computes a sequence of \emph{output states} from which a probability distribution over output characters can be computed:
\begin{align}
\svec\step{k} &= \dec(\svec\step{k-1}, \cvec\step{k}, \yvec\step{k-1}) \\
P(\yvec\step{k}) &= \softmax(\svec\step{k}).
\end{align}
\noindent The attention mechanisms produce the attention weights with the following computations, as in~\cite{luong2015effective}, with $\vvec\fst$, $\vvec\snd$, $\mathbf{W}_{\avec\fst}^s$, $\mathbf{W}_{\avec\snd}^s$, $\mathbf{W}_{\avec\fst}^h$, and $\mathbf{W}^h_{\avec\snd}$ being parameters to be learnt:
\begin{align}
\mathbf{\ascal}\fst\step{kn} &= \softmax (\vvec\fst \tanh (\left[ \mathbf{W}_{\avec\fst}^s \svec\step{k-1} ; \mathbf{W}_{\avec\fst}^h \hvec\fst\step{n} \right]) )\\
\mathbf{\ascal}\snd\step{km} &= \softmax (\vvec\snd \tanh (\left[ \mathbf{W}_{\avec\snd}^s \svec\step{k-1} ; \mathbf{W}_{\avec\snd}^h \hvec\snd\step{m} \right]) ).
\end{align}
Since both attention mechanisms provide context to the same decoder, we can tie the computation of the weights so that the two mechanisms share the $\vvec$ and $\mathbf{W}^s_{\avec}$ parameters . We refer to this version as \textit{tied} attention mechanism:
\begin{align}
\mathbf{\ascal}\fst\step{kn} &= \softmax (\vvec \tanh (\left[ \mathbf{W}_{\avec}^s \svec\step{k-1} ; \mathbf{W}_{\avec\fst}^h \hvec\fst\step{n} \right]) )\\
\mathbf{\ascal}\snd\step{km} &= \softmax (\vvec \tanh (\left[ \mathbf{W}_{\avec}^s \svec\step{k-1} ; \mathbf{W}_{\avec\snd}^h \hvec\snd\step{m} \right]) ).
\end{align}
\noindent If the two encoders share the same output size for their $\hvec\fst$ and $\hvec\snd$ vectors, then the two attentions could further share the $\mathbf{W}^h_{\ascal}$ parameters. This effectively merges them into one, \textit{shared} attention mechanism, so that:
\begin{align}
\mathbf{\ascal}\fst\step{kn} &= \softmax (\vvec \tanh (\left[ \mathbf{W}_{\avec}^s \svec\step{k-1} ; \mathbf{W}_{\avec}^h \hvec\fst\step{n} \right]) )\\
\mathbf{\ascal}\snd\step{km} &= \softmax (\vvec \tanh (\left[ \mathbf{W}_{\avec}^s \svec\step{k-1} ; \mathbf{W}_{\avec}^h \hvec\snd\step{m} \right]) ).
\end{align}
\noindent Furthermore, another line of work that is pertinent to our work is based in \textit{ensembling}. Traditionally, ensembles refer to models that have been trained on similar data for the similar task but their predictions are combined at inference time, usually achieving better performance than a single model. 

In our case, we explore ensembles of a transcription and a translation model. In traditional ensembling, the models are trained separately. Recently, however, \textit{coupled ensembles} were shown to outperform simple ensembles~\cite{dutt2017coupled}. In the \textit{coupled ensemble} setting (see Figure~\ref{fig:multisourcemodels}c), the two models are trained jointly, albeit they don't share any parameters. 

The two decoder outputs are averaged right before the softmax layer, in order to produce a single output probability distribution.
It was shown~\cite{dutt2017coupled} that this approach works better than combining the two predictions after the softmax layer: 
\begin{align}
\svec\fst\step{k} &= \deca(\svec\fst\step{k-1}, \cvec\fst\step{k}, \yvec\step{k-1}) \\
\svec\snd\step{k} &= \decb(\svec\snd\step{k-1}, \cvec\snd\step{k}, \yvec\step{k-1}) \\
P(\yvec\step{k}) &= \softmax(\frac{\svec\fst\step{k} + \svec\snd\step{k}}{2}).
\end{align}

\section{Related Work}

The \textit{speech translation} problem has been traditionally approached by feeding the output of a speech recognition system into a Machine Translation (MT) system. 
Speech recognition uncertainty was integrated with MT by using speech output lattices as input to translation models~\cite{ney1999speech,matusov2005integration}.
A sequence-to-sequence model for speech translation without transcriptions has been  introduced~\cite{duong-EtAl:2016:N16-1}, but was only evaluated on alignment.
Synthesized speech data were translated in~\cite{berard2016listen} using a model similar to the Listen Attend and Spell model~\cite{chan2016listen}, while a larger-scale study~\cite{berard2018end} used an end-to-end system for translating audio books between French and English.
Sequence-to-sequence models to both transcribe Spanish speech and translate it in English have also been proposed~\cite{weiss2017sequence}, by jointly training the two tasks in a multitask scenario with two decoders sharing the speech encoder. This model was extended by us~\cite{anastasopoulos+chiang:naacl2018}, with the translation decoder receiving information both from the speech encoder and the transcription decoder.

Multi-source models have also been studied, for statistical MT~\cite{och2001statistical} and neural MT~\cite{zoph2016multisource}. The two inputs are the same sentence in two languages, and the model is trained on tri-lingual text. Multi-source ensembles for MT have also been explored~\cite{garmash2016ensemble}, where the two ensemble components were trained separately, using text input in different languages.

The only previous work that operates under similar assumptions to ours, that is, having access only to translations of the train and test utterances and no other parallel data, is \texttt{LatticeTM}~\cite{adams2016learning}, a model that composes word lattices (the result of ASR) with the weighted finite-state transducer of a translation model that is jointly learned. They showed consistent reductions in word error rate (WER) on two datasets, including CALLHOME.
Note that since they use word lattices to represent the speech recognition output and do not attempt acoustic modeling, their setting is easier than ours. 
Our approach, instead, operates directly on the speech signal.

\section{Experiments}

\subsection{Implementation}
Our models are implemented in DyNet~\cite{neubig2017dynet}.\footnote{Our code is available online at \url{https://bitbucket.org/antonis/dynet-multisource-models}.} We use a dropout of 0.2, and train using Adam with initial learning rate of $0.0002$ for up to 300 epochs. The hidden layer and the attention size are 512 units.  The acoustic encoder employs a 3-layer speech encoding scheme~\cite{duong-EtAl:2016:N16-1}. The first bidirectional layer receives the audio sequence in the form of 39-dimensional Perceptual Linear Predictive (PLP) features~\cite{hermansky1990perceptual} computed over overlapping 25ms-wide windows every 10ms. The second and third layers consist of LSTMs with hidden state sizes of 128 and 512 respectively. Each layer encodes every second output of the previous layer. The translation encoder uses a bi-directional LSTM layer. The input and output character embedding size is set to 32.

For testing, we select the model with the best performance on dev. At inference time, we use a beam size of 4 and the beam score includes length normalization~\cite{wu2016google} with a weight of 0.8, which has been found to work well for low-resource neural Machine Translation~\cite{nguyen2017transfer}.

\subsection{Data}
We evaluate all our models on three diverse datasets.

\emph{Mboshi-French}:
Mboshi (Bantu C25 in the Guthrie classification) is a language spoken in Congo-Brazzaville, without standard orthography. We use a corpus of 5517 parallel utterances (about~4.4 hours of audio) collected from three native speakers using the LIG-Aikuma app for the BULB project~\cite{adda2016breaking,godard2017very}. The corpus provides non-standard grapheme transcriptions (produced by linguists to be close to the language phonology) 
as well as French translations. We sampled~100 segments from the training set to be our dev set, and used the original dev set (514 utterances) as our test set. 
\begin{table}[t]
    \centering
    \caption{Multi-Source models achieve lower Character Error Rates (CER) on all three target languages, even in extremely low resource settings (Ainu, Mboshi). In Spanish, we observe an~8.4\% reduction in CER.}
    \label{tab:cer_results}
    \begin{tabular}{c|ccc}
    \toprule
        \multirow{2}{*}{Source} & \multicolumn{3}{c}{Transcription CER} \\
          & Ainu & Mboshi & Spanish \\
    \midrule
        speech (audio)  & 40.7 & 29.8 & 52.0\\
        translation (text) & 74.9 & 68.2 & 44.6\\
    \midrule
        coupled ensemble & \textbf{40.6} & 36.8 & 42.2\\
        multi-source & 46.0 & 37.5 & 41.6\\
        +tied & 41.4 & 32.6 & \textbf{37.6} \\
        +shared & \textbf{40.6} & \textbf{28.6} & 38.7\\
    \bottomrule     
    \end{tabular}
\end{table}

\emph{Ainu-English}:
Hokkaido Ainu is the sole surviving member of the Ainu language family and is generally considered a language isolate. As of 2007, only ten native speakers were alive. The Glossed Audio Corpus of Ainu Folklore 
provides 10 narratives (about~2.5 hours of audio), transcribed at the utterance level, glossed, and translated in Japanese and English.\footnote{\texttt{http://ainucorpus.ninjal.ac.jp/corpus/en/}} All narratives were collected from the same speaker; the audio quality, though, is quite low, as the recordings were performed in an often noisy environment. Furthermore, the narratives have a distinct prosodic quality to them: at least two of them are more sung than narrated, rendering the dataset even harder to work with. This is further addressed in Section~\ref{sec:results}.

Since there does not exist a standard train-dev-test split, we employ a cross validation scheme for evaluation purposes. In each fold, one of the 10 narratives becomes the test set, with the previous one (mod 10) becoming the dev set, and the remaining 8 narratives becoming the training set. The models for each of the~10 folds are trained and tested separately. On average, for each fold, we train on about~2000 utterances; the dev and test sets consist of about~$270$ utterances. In Section~\ref{sec:results} we report results on the concatenation of all folds, 
but also provide a breakdown of the performance in each fold.

\emph{Spanish-English}: 
Spanish is obviously neither an endangered nor a low-resource language, but we pretend that it is one, by not making use of any Spanish resources like language models or pronunciation lexicons. We use the Spanish CALLHOME corpus (LDC96S35), which consists of telephone conversations between Spanish native speakers based in the US and their relatives abroad (about~20 total hours of audio) with more than 240 speakers. We use translations created by \cite{post2013improved} and keep the original train-dev-test split, with the training set comprised of 80 conversations and dev and test of 20 conversations each. Unlike the other two datasets, there is no speaker overlap between the train and the test set.

\begin{table}[t]
    \centering
    \caption{Evaluating the output with word-level BLEU, multi-source models significantly improve upon the baselines on higher-resource settings (Spanish). On Ainu, the best model performs on par with the very competitive baseline.}
    \label{tab:bleu_results}
    \begin{tabular}{c|cc}
    \toprule
        \multirow{2}{*}{Source} & \multicolumn{2}{c}{Transcription BLEU} \\
          & Ainu & Spanish \\
    \midrule
        speech (audio)  & \textbf{28.92} & 9.41 \\
        translation (text) & 5.89 & 14.73\\
    \midrule
        coupled ensemble & 26.99 & 16.94 \\
        multi-source & 24.03 & 17.59\\
        +tied & 26.95 & \textbf{20.82}\\
        +shared & \textbf{28.57} & 19.47\\
    \bottomrule     
    \end{tabular}
\end{table}

\begin{figure}

\pgfplotstableread[row sep=\\,col sep=&]{
story & base & best \\
1 & 36.0351915016 & 33.8821655679 \\
2& 35.7105543537 & 38.2103198929 \\
3& 31.4784769833 & 32.0486059302 \\
4& 27.921633991 & 28.074789417 \\
5& 34.9436346898 & 34.8969681394 \\
6& 91.2408190501 & 81.3695052681 \\
7& 67.0358483597 & 62.7334199139 \\ 
8& 27.5434049745 & 26.7265385274 \\
9& 39.7143557591 & 40.0810785502 \\
10& 45.151644942 & 43.3646013018 \\
}\ainudata

\definecolor{bblue}{HTML}{4F81BD}
\definecolor{rred}{HTML}{C0504D}

\begin{tikzpicture}
    \begin{axis}[
            ybar,
            bar width=.2cm,
            width=8.33cm,
            height=6cm,
            ymajorgrids=true,
            yminorgrids=true,
            legend style={at={(0.8,0.95)},
                anchor=north,legend columns=1},
            xtick=data,
            tick pos=left,
            ymin=0,ymax=100,
            ylabel={CER},
            ylabel near ticks,
            xlabel={Ainu Narrative}
        ]
        \addplot [style={black,postaction={pattern=north east lines},fill=white,mark=none}] table[x=story,y=base]{\ainudata};
        \addplot [style={black,fill=black,mark=none}] table[x=story,y=best]{\ainudata};
        \legend{baseline, multisource}
    \end{axis}
\end{tikzpicture}
\caption{Character Error Rates of the best baseline system and our best multisource system for each Ainu narrative. The gains of using the translations are apparent in the cases that are harder for a speech-only system: narratives 6 and 7 are more sung than narrated, rendering them harder to transcribe.}
\label{fig:ainu}
\end{figure}
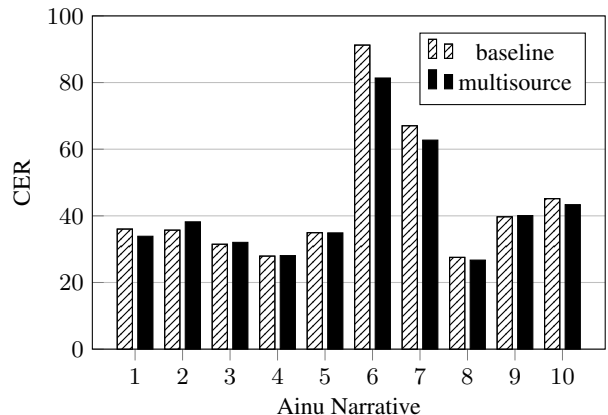
\begin{figure*}[t]
\centering
\begin{tabular}{c|cc}
    \multicolumn{1}{r|}{\includegraphics[width=0.285\textwidth, height=1.6em]{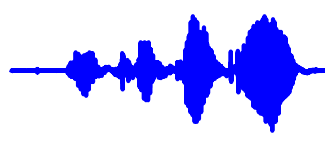}}
    &
    \multicolumn{1}{r}{\includegraphics[width=0.285\textwidth, height=1.6em]{figures/speech-speechexample.png}}
    & 
    \multirow{2}{*}[0.75em]{\includegraphics[width=0.229\textwidth]{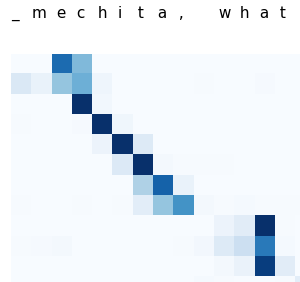}}\\
    \includegraphics[width=0.3\textwidth]{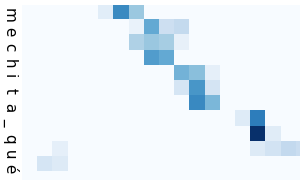}
    &
    \includegraphics[width=0.3\textwidth]{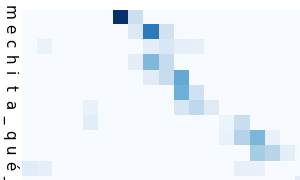}
    &\\
    (a) speech-source & (b) multi-source: speech & (c) multi-source: translation
\end{tabular}
\caption{Attentions on a speech sample from the dev set, that includes a proper name (``\texttt{Mechita}") unseen during training. The multi-source model (using a \textit{shared} attention mechanism) receives informative context from the translation so as to produce the output.}
\label{fig:attentions}
\end{figure*}

\subsection{Results}
\label{sec:results}

The character error rates of the baseline and our multi-source models are presented in Table~\ref{tab:cer_results}. In the two extremely low-resource datasets, Ainu and Mboshi, we find that the speech-only baselines are quite competitive. The very small number of speakers (one and three, respectively) makes the speech transcription task easier. In contrast, the translation task is much harder with so little data, as is also confirmed by the poor performance of the translation-only models in Ainu and Mboshi. 

In the relatively higher-resource CALLHOME dataset, the translation-only model outperforms the speech-only one, with an improvement of~7.4 points in CER, a~12.3\% reduction. This is most likely due to the lack of speaker overlap across the training and the test set. In this setting, the acoustic modelling part is harder than encoding the translation sentence.

For completeness, in Table~\ref{tab:bleu_results} we also report word-level BLEU scores~\cite{papineni2002bleu}, the most common  evaluation metric for Machine Translation. A higher BLEU almost always translates to lower WER. We only report results on Ainu and Spanish as Mboshi does not have standardized word segmentation rules. The BLEU scores reinforce our previous analysis. In the Ainu dataset, where a translation-only model is impossible to train (as outlined by the BLEU score of about~5.9), the multi-source model with \textit{shared} attention mechanism performs on par with the speech-only model, as does the coupled ensemble model. In the CALLHOME dataset, our multi-source model with the \textit{tied} attention mechanism achieves significantly higher BLEU scores than the translation-only or the coupled ensemble models. 
In addition, our best model's WER on the CALLHOME test set is~53.0, outperforming \texttt{LatticeTM} that achieved a WER of~56.2, despite the fact that our model operates in a harder setting, trained and tested directly on speech.

The performance of each fold of cross-validation for Ainu is shown in Figure~\ref{fig:ainu}. For each narrative, it compares the speech-only baseline system
with our best multi-source system. The overall performance of the speech-only single-source model and our best model is similar with a CER of~40.7 and~40.6 respectively. A possible reason is that all the Ainu stories are narrated by the same speaker, making it a generally easier task for a speech recognition system. 
But we also see that in the cases where speech transcription is harder, translation information does help. Namely, narratives~6 and~7 are \textit{sung}, making them  harder to transcribe with a speech-only system trained on spoken data, as indicated by the higher error rates:~91.2 and ~67.0, respectively. The multi-source models achieve noticeable improvements of~9.9 and~4.3 points on these narratives.

We further quantify the effect of the different sharing mechanisms for the attentions. Using word-level forced alignments on the CALLHOME dataset ~\cite{duong-EtAl:2016:N16-1} we can evaluate the accuracy of the attention.
Treating the forced alignments as reference, we compute the percentage of the weights of the attention over the speech source that fall within the boundaries of the forced alignment spans. 
Note that the forced alignments naturally include noise, so they should be treated as a ``silver standard." 
However, they can still provide indications that could reveal the effect of parameter sharing.

We computed the average sum of this \textit{attention accuracy} by forced decoding on the CALLHOME development set. 
We find that the average sum for the speech single-source model is almost~71\%, a value similar to the average sums of the attention accuracy of the coupled ensemble and the multi-source model that employs no sharing mechanism. 
Instead, the attention accuracy of the model with the shared mechanism is almost~75\%. The model with tied attentions, which achieves the best results on CALLHOME, has an attention accuracy of~76\%.

Figure~\ref{fig:attentions} presents the attention weights over a sample taken from the development set, produced by forced decoding. The segment includes an out-of-vocabulary word, the name \emph{Mechita}, never seen during training. The attention weights over the speech source with the single-source model (3a) are not too different from the weights of the multi-source model with tied attentions (3b). However, the multi-source model in this case takes advantage of the translation and receives most of its context from the text source (3c), as the attention weights over the characters of the name are quite high (albeit, off-by-one, as often is the case in neural attention-based translation).

\section{Conclusion}

We presented multi-source neural architectures that receive an audio segment and its translation and produce a character-level transcription in low-resource settings.
We showed that providing the translation as an additional input signal is beneficial to the transcription task, as our models outperform the single-source baselines. Furthermore, we find that sharing the decoder and the attention parameters leads to lower character error rate over either a coupled ensemble architecture or simple attention mechanisms without parameter sharing.

These results will hopefully lead to new tools for endangered language documentation. 
Projects like the BULB project that aim to collect about $100$ hours of audio with translations, stand to benefit from our approach, since it would be impractical to manually transcribe this much audio for many languages. 
We hope that this work will provide a concrete basis for leveraging translations in a language documentation pipeline.

\clearpage

\bibliographystyle{IEEEtran}
\bibliography{References}

\end{document}